\documentclass[letterpaper, 10 pt, conference]{ieeeconf}  %
\usepackage[utf8]{inputenc}
\usepackage{caption}
\usepackage{cite}
\usepackage{url}

\newcommand{\etal}{\textit{et al.}}

\IEEEoverridecommandlockouts                              %

\title{\LARGE \bf
Structured Domain Randomization: Bridging the Reality Gap by Context-Aware Synthetic Data
}

\author{Aayush Prakash\thanks{Authors are affiliated with NVIDIA.  Email: {\tt\footnotesize \{aayushp, sboochoon, markb, dacunamarrer, ecameracci, gstate, oshapira, sbirchfield\}@nvidia.com}}%
 \\ Eric Cameracci \and Shaad Boochoon \\ Gavriel State \and Mark Brophy \\ Omer Shapira  \and David Acuna \\ Stan Birchfield}

\usepackage{times}
\usepackage{epsfig}
\usepackage{graphicx}
\usepackage{amsmath}
\usepackage{amssymb}
\usepackage{bbm}  %
\usepackage{comment}
\usepackage{subfigure}

\usepackage[ruled,linesnumbered]{algorithm2e}
\usepackage{booktabs}
\usepackage{array, makecell, tabularx}
\usepackage{xspace}

\usepackage{subfigure}
\usepackage{xcolor}
\usepackage{rotating}
\usepackage[export]{adjustbox}
\usepackage{mathtools}

\newcommand{\bfc}{\ensuremath{{\bf{c}}}}

\newcommand{\bfg}{\ensuremath{{\bf{g}}}}

\newcommand{\bfo}{\ensuremath{{\bf{o}}}}

\begin{document}

\maketitle

\thispagestyle{empty}
\pagestyle{empty}

\begin{abstract}

We present structured domain randomization (SDR), a variant of domain randomization (DR) that takes into account the structure and context of the scene.
In contrast to DR, which places objects and distractors randomly according to a uniform probability distribution, SDR places objects and distractors randomly according to probability distributions that arise from the specific problem at hand.
In this manner, SDR-generated imagery enables the neural network to take the context around an object into consideration during detection.  
We demonstrate the power of SDR for the problem of 2D bounding box car detection, achieving competitive results on real data after training only on synthetic data.
On the KITTI easy, moderate, and hard tasks, we show that SDR outperforms other approaches to generating synthetic data (VKITTI, Sim~200k, or DR), as well as real data collected in a different domain (BDD100K).  Moreover, synthetic SDR data combined with real KITTI data outperforms real KITTI data alone.

\end{abstract}

\section{Introduction}
Training deep networks for computer vision tasks typically requires large amounts of labeled training data.
Annotating such data is laborious and time-consuming, thus making it cost-prohibitive for tasks for which the labels are particularly difficult to acquire, such as instance segmentation, optical flow estimation, or depth estimation.  
Even for problems like 2D bounding box detection, there is a motivation to avoid the expensive labeling process.

Synthetic data is an attractive alternative because data annotation is essentially free.
Recently a number of synthetic datasets \cite{butler2012eccv,handa2015arx:sn,DFIB15iccv,mayer2015arx,qiu2016arx:uncv,zhang2016arx:unst, mccormac2016arx:snrgbd,ros2016cvpr:syn,Richter_2016_ECCV,gaidon2016CVPR,Mueller2017ue4,Tsirikoglou2017exporter} have been generated for training deep networks.
These datasets require either carefully designed simulation environments or the existence of annotated real data as a starting point.
To alleviate such difficulties, Domain Randomization (DR)~\cite{tobin17:dr,dr_tremblay_2018} proposes to randomize the input so as to minimize the need for artistic design of the environment or prior real data.
Recent work~\cite{dr_tremblay_2018} demonstrated the ability of DR to achieve state-of-the-art in 2D bounding box detection of cars in the KITTI dataset~\cite{Geiger2012CVPR}.  However, the results of that research were limited to larger objects (KITTI Easy) for which sufficient pixels exist within the bounding box for neural networks to make a decision without the surrounding context.

In this paper we extend that work to handle more challenging criteria (KITTI Moderate and Hard). The ground truth of these criteria include small, occluded (partial to heavy), and significantly truncated objects. In these cases, such objects occupy only a few pixels in the image, thus making it necessary to take into account the surrounding context of the scene.  To address this problem, we propose \emph{structured domain randomization (SDR)}, which adds structure and context to domain randomization (DR).  
We present a methodology to train deep networks for object detection using only synthetic data generated by SDR, and we show that the results from this process not only outperform other approaches to generating synthetic data, but also real data from a different domain.\footnote{Video is at \url{https://youtu.be/1WdjWJYx9AY}.}
Our contributions are as follows:
\begin{itemize}
	\item We introduce a context-aware domain randomization procedure called structured domain randomization (SDR).
	We describe an implementation of SDR for object detection that takes into account the structure of the scene when randomly placing objects for data generation, which enables the neural network to learn to utilize context when detecting objects.
	\item We demonstrate that SDR achieves state-of-the-art for 2D object detection on KITTI (easy, moderate, and hard).  The performance achieved is better than both virtual KITTI (VKITTI)~\cite{gaidon2016CVPR} and GTA-based Sim~200k~\cite{johnson2017icra:ditmatrix}, the two most commonly used synthetic datasets for object detection.  Performance is also better than real data, BDD100K~\cite{bdd100k}, from a different domain.
\end{itemize}

\section{Related Work}
\label{sc:related}

Synthetic data has been used in a myriad of vision tasks where labeling images ranges from 
tedious to borderline impossible. Applications like optical flow~\cite{DFIB15iccv,butler2012eccv},
scene flow~\cite{mayer2015arx},
classification~\cite{borrego2018arxiv},
stereo~\cite{qiu2016arx:uncv,zhang2016arx:unst}, 
semantic segmentation~\cite{Richter_2016_ECCV, ros2016cvpr:syn},
3D keypoint extraction~\cite{Suwajanakorn2018nips:latent},
object pose~\cite{Mueller2017ue4} and
3D reconstruction ~\cite{handa2015arx:sn, mccormac2016arx:snrgbd} have all
benefited from the use of synthetic training data.

Synthetic data has also been utilized for object detection, the problem for which our SDR 
algorithm proposes an approach.
Gaidon \etal~\cite{gaidon2016CVPR} 
created a synthetic clone of five videos from the KITTI detection dataset called Virtual KITTI (VKITTI).
They demonstrated the ability to
train a model for object detection using synthetic data that learns features which are generalizable to real images.
Johnson-Roberson \etal~\cite{johnson2017icra:ditmatrix} used synthetic data 
captured from GTA V to train an object detector for cars and demonstrated that
photo-realism aided in the training of DNNs. 
Mueller \etal~\cite{Mueller2017ue4} built Sim4CV for autonomous navigation and tracking 
on top of Unreal Engine, focusing on generating realistic synthetic scenes like 
those seen in ~\cite{johnson2017icra:ditmatrix} but further studying problems of control and
autonomous navigation when utilizing synthetic data.

Hinterstoisser \etal~\cite{hinterstoisser2017arx:pretrain} eschewed the idea of photorealism and 
generated images by adding Gaussian noise to the foreground of the rendered image and performing 
Gaussian blurring on the object edges to better integrate with the background image.
The resulting synthetic data was used to train the later layers of a neural network while freezing the early layers pretrained on real data (\emph{e.g.}, ImageNet).
Dwibedi \etal~\cite{dwibedi2017iccv:cutpaste} extended this idea, using 
multiple blending techniques and 
varying the blending parameters to make the detector more robust to object boundaries 
and thus improve performance.
Mayer \etal~\cite{Mayer2018ijcv:synthetic} found that when training neural networks 
to perform optical flow estimation, simplistic data with augmentation is sufficient
(they conclude: ``realism is overrated") but concede that the same may not be true for 
high-level tasks like object recognition. Indeed, our work offers further evidence in 
this direction.

Tobin \etal~\cite{tobin17:dr} introduced the concept of Domain Randomization (DR), in which realistic rendering is avoided in favor of random variation. 
Their approach randomly varies the texture 
and color of the foreground object, the background image, the number of lights in the 
scene, the pose of the lights, the camera position, and the foreground objects. The goal is to
close the reality gap by generating synthetic data with sufficient variation that the network views real-world data as just another variation.
Using DR, they trained a neural network to estimate the 3D world position of various shape-based objects with respect to a robotic arm fixed to a table. Sundermeyer \etal~\cite{Sundermeyer2018eccv:implicit} use DR for object detection and 6D pose estimation,  achieving competitive results compared to pose estimation using real data.

Our previous work in~\cite{dr_tremblay_2018} used DR to train an object detector for cars, which was tested on KITTI, similar to the work presented here.
In that research, we learned that DR requires a large amount of data to train given the amount of variation, often the network finds it hard to learn the correct features, and the lack of context prevents DR from detecting small vehicles.
The research described in this paper aims to address these limitations.
Other researchers~\cite{ZhuSegDeepM15} have also found context 
to be important. Georgakis \etal~\cite{Georgakis2017boxsyn} create training data for indoor robotics by 
locating planes in background images and pasting foreground objects onto them, an acknowledgment of 
the importance of context.

\section{Structured Domain Randomization (SDR)}

Structured Domain Randomization (SDR) is a general technique for procedurally generating synthetic random images that preserve the structure, or context, of the problem at hand.  
In our formulation, SDR involves three types of components:  1) global parameters, 2) one or more context splines, and 3) objects that are placed along the splines.  

\begin{figure}
\center
\includegraphics[width=0.25\textwidth]{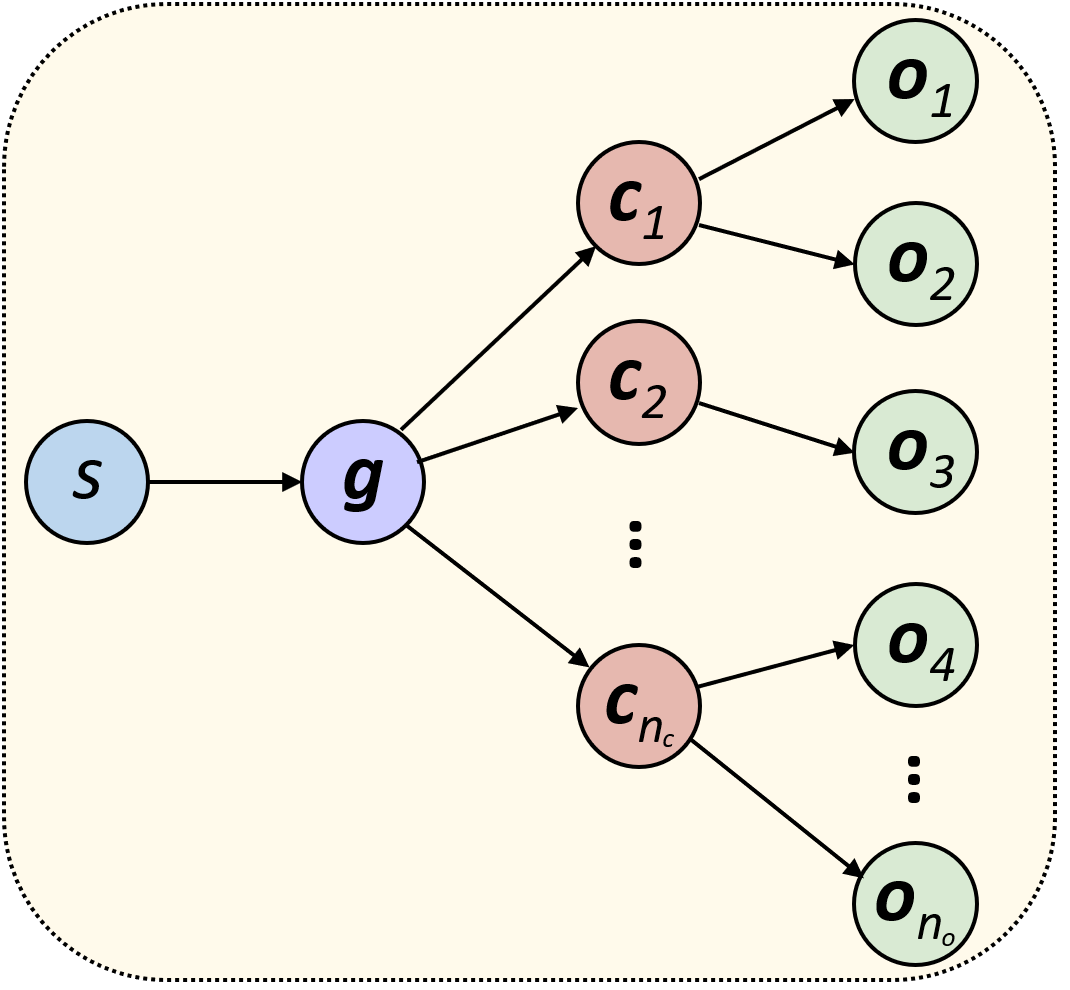}
\caption{Probabilistic relationship among different components of SDR. The scenario ($s$) determines the global parameters ($\bfg$), which govern the context splines ($\bfc_i$), upon which the objects ($\bfo_j$) are placed.  The context splines capture the structure of the scene.  The image is rendered from these parameters, splines, and objects.}
\label{fig:graphical_model}
\end{figure}

The joint probability of generating a particular image $I$ and the parameters, splines, and objects is given by the following:
\begin{align}
p(I,s,\bfg,\bfo_{1..n_o},\bfc_{1..n_c}) &=
p(I|s,\bfg,\bfo_{1..n_o},\bfc_{1..n_c}) \nonumber \\
& \hspace{-4em} \cdot \prod_{j=1}^{n_o} p(\bfo_j | \bfc_i) \prod_{i=1}^{n_c} p(\bfc_i | \bfg) p(\bfg | s) p(s),
\end{align}
which is depicted in Fig.~\ref{fig:graphical_model}.  

\begin{figure*}
    \centering
    \begin{tabular}{ccc}
			\includegraphics[width=0.3\textwidth]{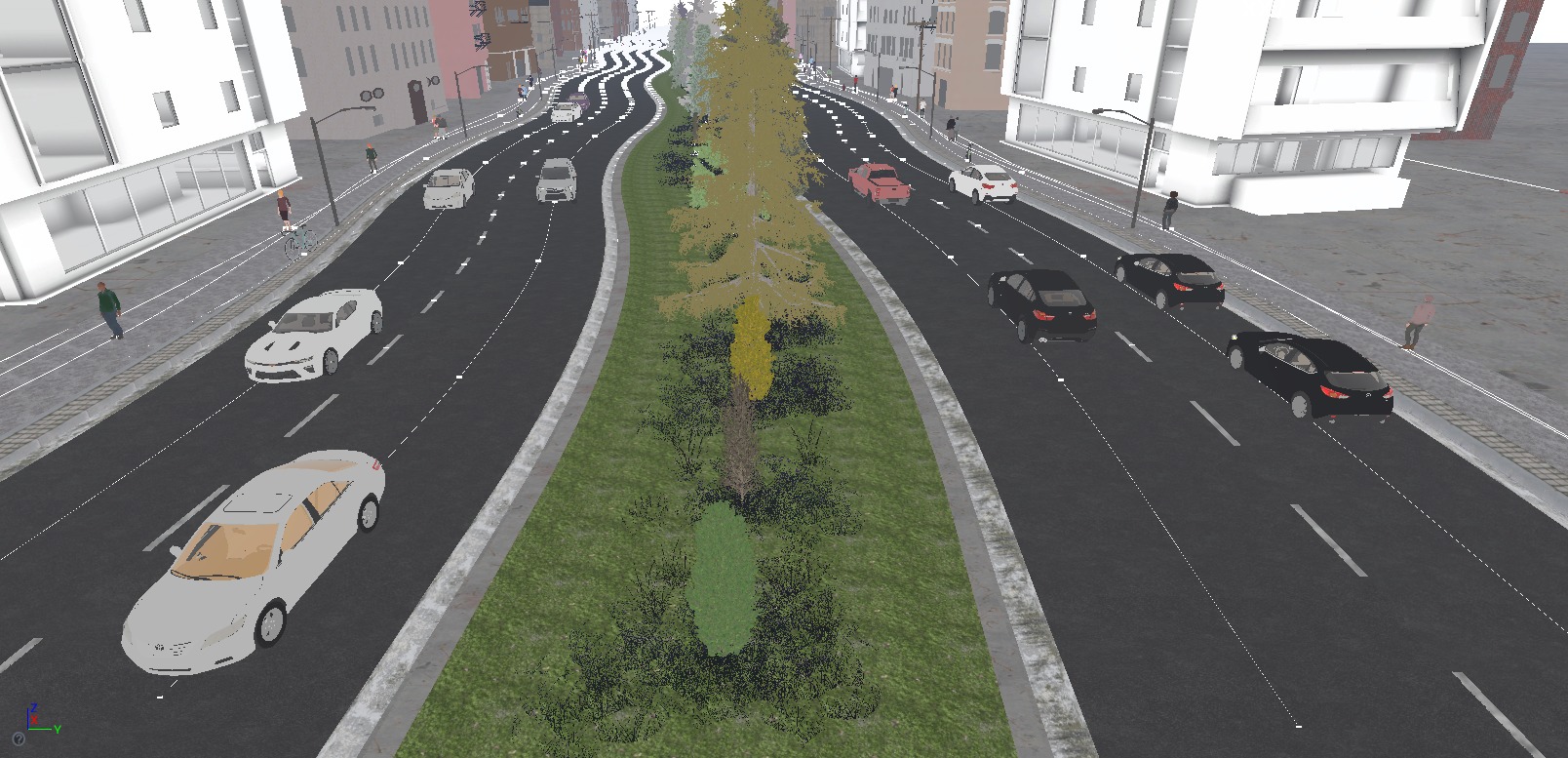} &
			\includegraphics[width=0.3\textwidth]{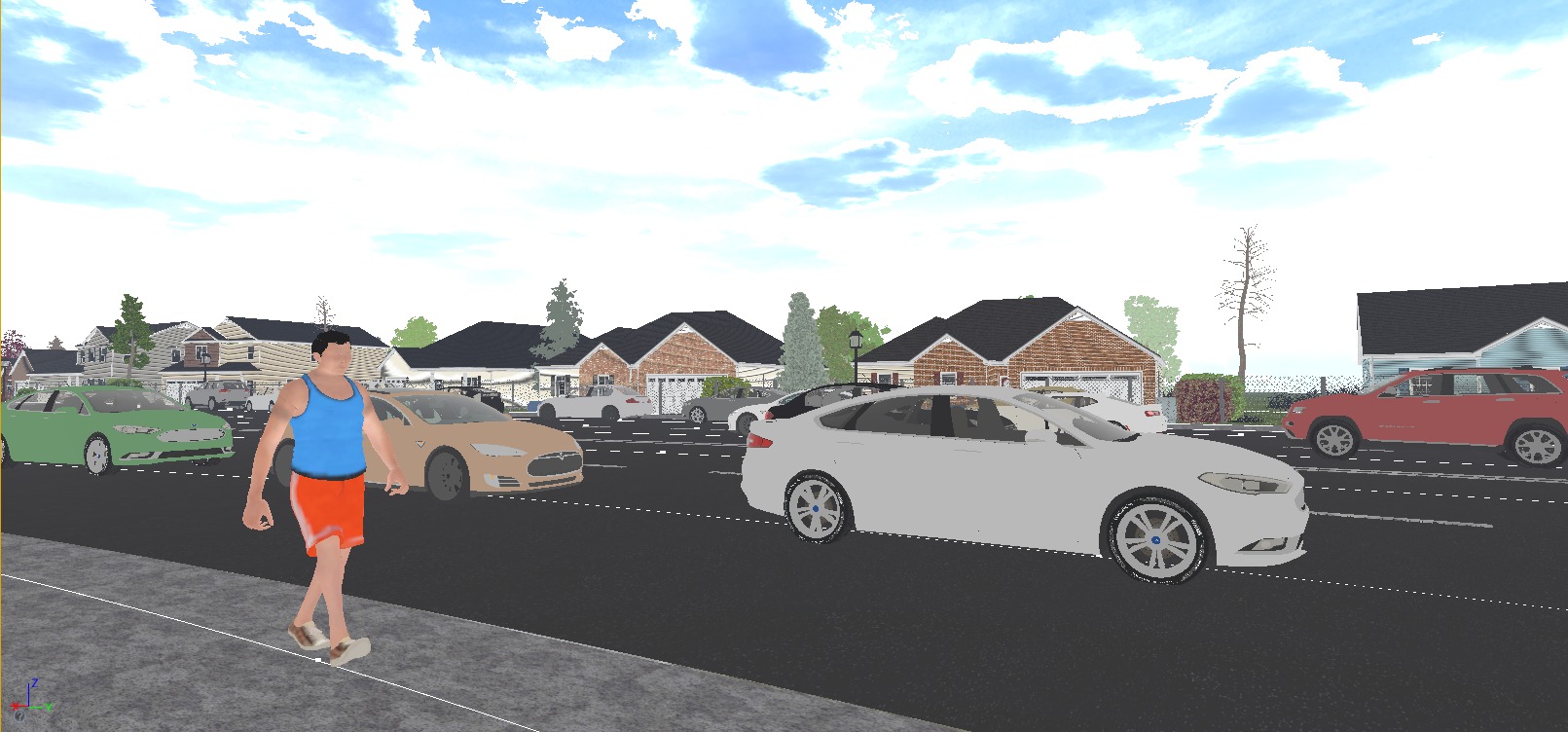}  &
      \includegraphics[width=0.3\textwidth]{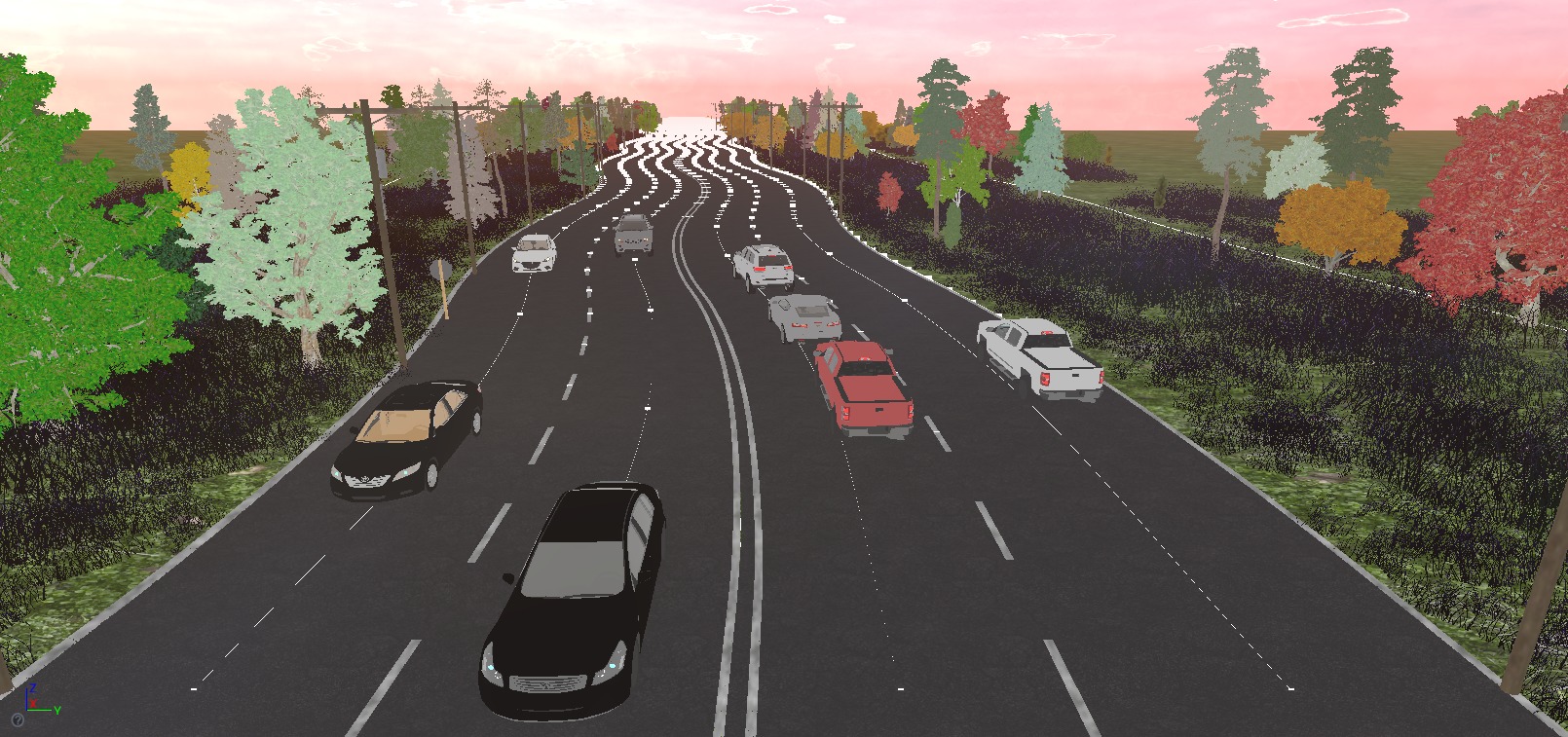} \\
			
			 \includegraphics[width=0.3\textwidth]{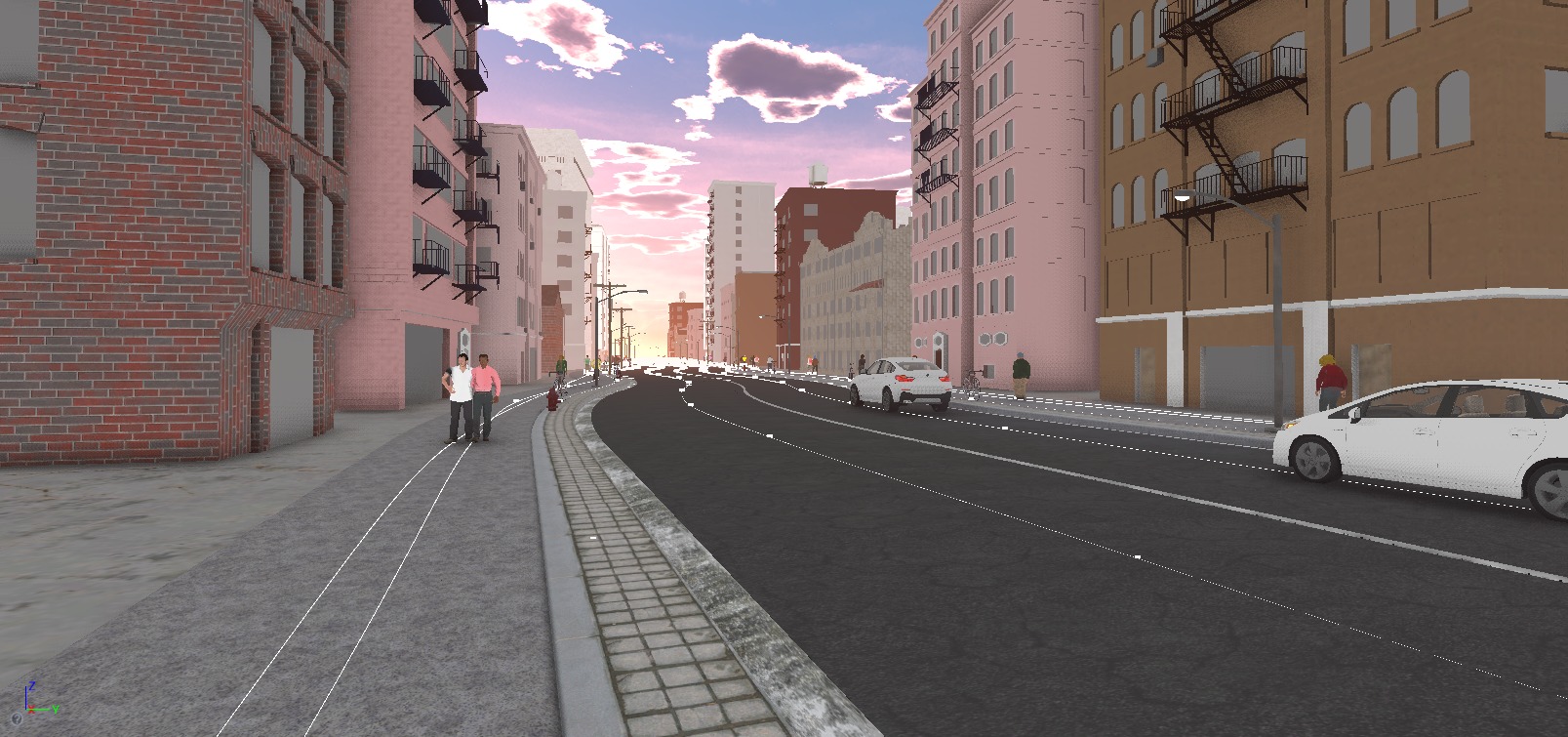} &
			\includegraphics[width=0.3\textwidth]{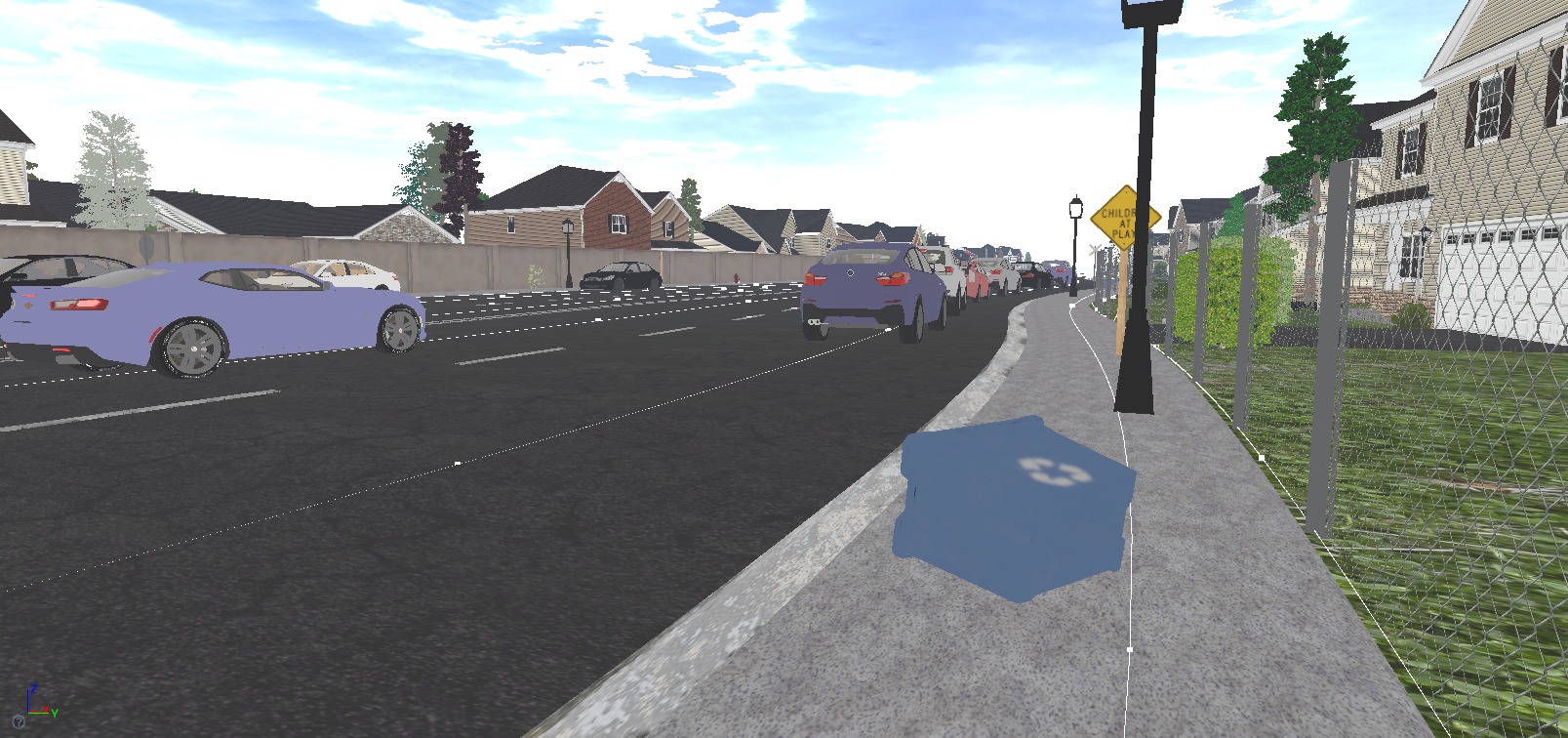}  &
		  \includegraphics[width=0.3\textwidth]{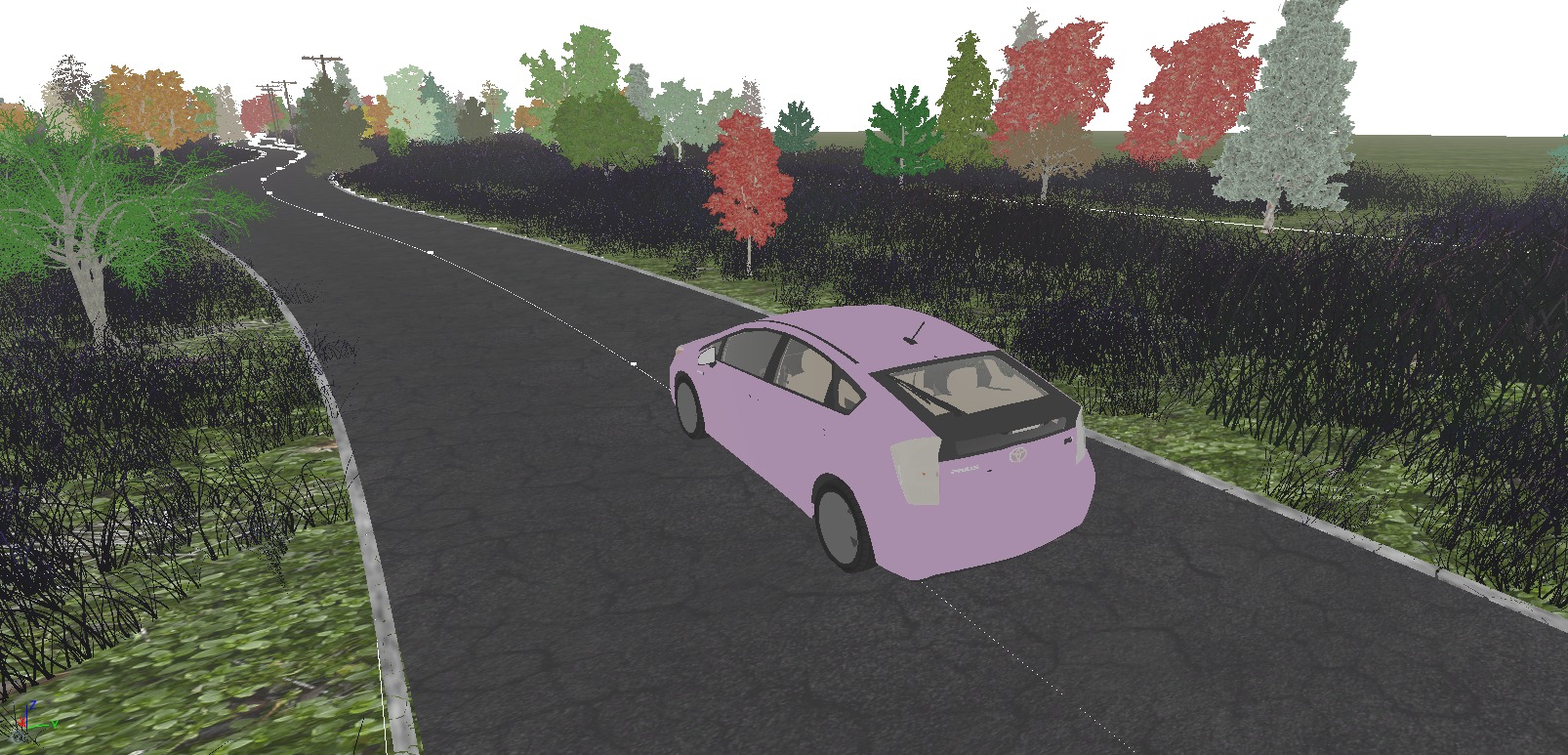} \\
			
			 \includegraphics[width=0.3\textwidth]{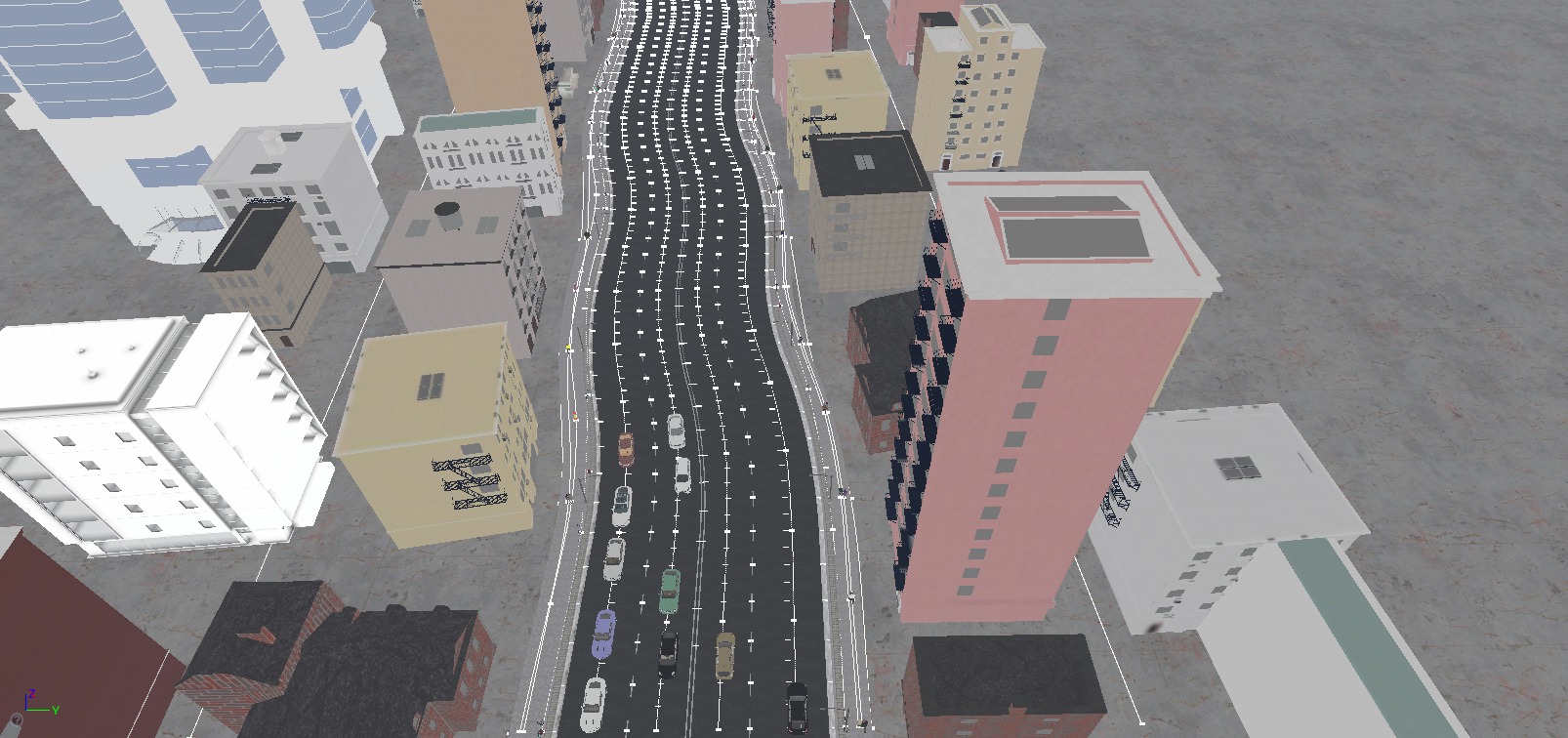} &
			\includegraphics[width=0.3\textwidth]{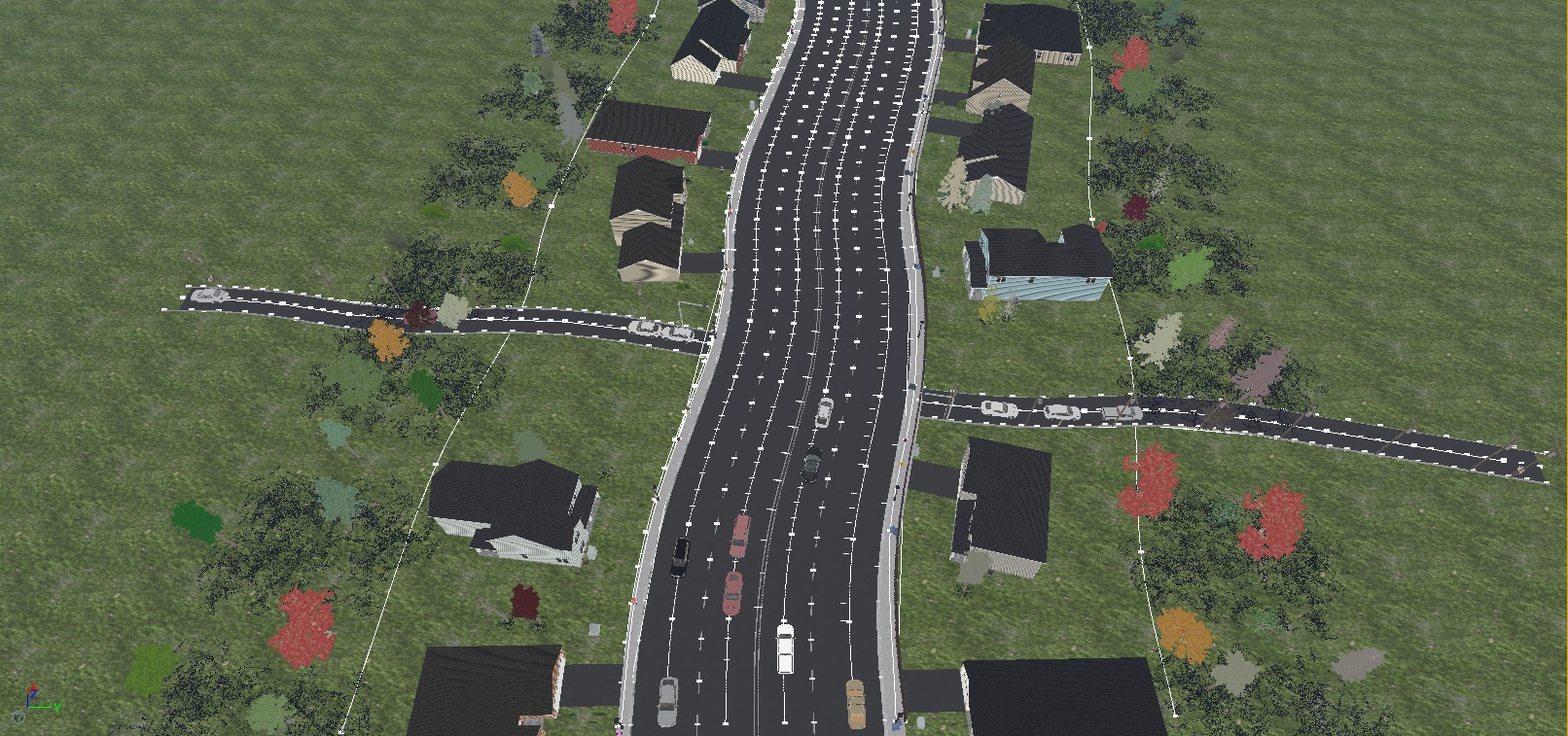}  &
		  \includegraphics[width=0.3\textwidth]{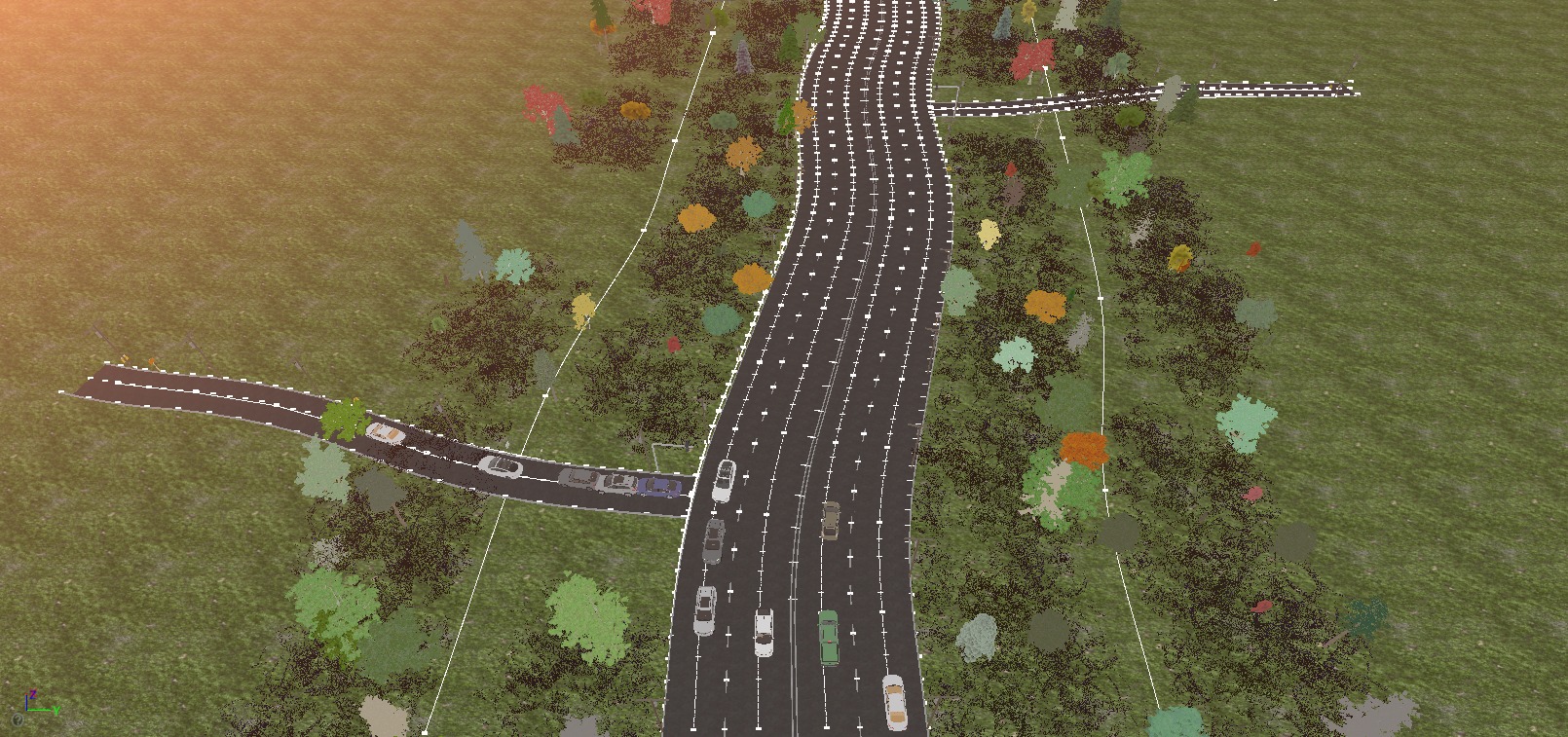} \\

			Urban & Suburban & Rural
			
 \end{tabular}
    \caption{In Structured Domain Randomization (SDR), a scenario is chosen at random, then global parameters (road curvature, lighting, camera pose, etc.), which cause context splines (road lanes, sidewalks, etc.) to be generated, upon which objects (cars, trucks, pedestrians, cyclists, houses, buildings, etc.) are placed.  The context splines are shown as thin overlaid white lines with white dots indicating control points. Note that these illustrative images were generated from camera viewpoints different from those used for training.}. 
    \label{fig:method}
\end{figure*}

First, a scenario $s$ is determined randomly.  In our implementation, there are approximately 20 scenarios, such as ``rural 2-lane road'', ``suburban 4-lane road with a sidewalk'', or ``urban 6-lane road with a grassy median and a sidewalk''.  The scenario is chosen from a uniform distribution across all possibilities. Figure~\ref{fig:method} shows examples of some scenarios.

Once the scenario has been chosen, the global parameters $\bfg$ are determined.  These include the spline shape, which is specified by $n_s=100$ control points, with a random right/left/straight decision made after every fixed subset of control points.  Each right/left turn is fixed at 30 degrees and only allowed if the road is already heading the opposite direction (so as to avoid hairpin turns).  Other global parameters include the azimuth/elevation of the sun, time of day, the color temperature and intensity of the sun, sky color, cloud density/positions, the camera yaw/pitch/FOV, the maximum number of vehicles per lane, and so forth.  The global parameters also include the number of lanes, whether a median exists, when a sidewalk exists, and so forth.

The parameters of each context spline $\bfc_i$ are determined by the global parameters.  The $n_c$ context splines are adjacent to one another and share their shape.  There is one context spline for each lane, one for the median, one for each sidewalk, one for each gutter, and one for each side stretch.  These splines receive random colors and textures that govern their appearance, such as the type of grass, darkness of the asphalt, and type of concrete.  Overlaid on these splines are various imperfections, such as potholes, cracks, and oil spills on the road.  
Figure~\ref{fig:method} shows these different type of context splines $\bfc_i$ as white lines with control points.

The $n_o$ objects $\bfo_j$ are placed randomly on the context splines. We associate different kind of objects with different kind of context splines. Lane splines receive vehicles, sidewalks receive pedestrians and cyclists, side stretch splines receive buildings, houses, and street signs, and so forth.  Vehicles are placed in lanes as follows:  First, a maximum number of vehicles is determined randomly for each lane, up to some global maximum.  Within each lane, a single vehicle is placed at a random distance from the observing vehicle, near the center of the lane and aligned with the road direction.  The second vehicle is placed randomly in the road with a minimum offset distance between it and the first vehicle.  This process continues until either the maximum number of vehicles for that lane, or the total maximum number of vehicles for the image, has been reached.  Similar procedures govern the placement of pedestrians, cyclists, buildings, and road signs.
Figure~\ref{fig:method} shows these objects $\bfo_j$ placed on the white context splines $\bfc_i$.

By contrast, with DR, the probability of an image $I$ being generated is not dependent on context. Rather, the objects are placed randomly in the scene with backgrounds from image datasets such as COCO\cite{Lin2014COCO} or ImageNet\cite{imagenet_cvpr09}. DR lacks the structure present in SDR, that is, it does not have the conditional dependence $p(\bfo_j|\bfc_i)$ of the objects on the context.

The procedure is implemented using the scene generator of the UE4 game engine.\footnote{\url{https://www.unrealengine.com/}}
The scene generator uses an exporter to generate labels for supervised learning.
The exporter reads various render buffers from UE4 (e.g., lit, depth, stencil) and saves them as image files representing the rendered scene, ground truth depth, and segmentation mask.
The exporter also uses the vertex data of 3D object meshes to generate 2D bounding boxes, object-oriented or axis-aligned 3D bounding boxes, truncation, and occlusion values. 

Our implementation of SDR includes 74 car models, 13 truck models, 5 bicycle models, 41 building models, 87 house models, 24 tree models, 20 pedestrian models, and 100 road sign models.  Other models included are street lights, walls, fences, fire hydrants, recycling bins, telephone poles, traffic lights, and utility boxes---with a small number (1--3) of each.
For DR data, we used these same models as distractors and the same cars as objects of interest.

We use Substance\footnote{\url{https://www.allegorithmic.com/substance}} to randomize the materials of both object and context splines. These include the paint of vehicles based on 9 standard colors, lightness variation, roughness, and metallic properties. For DR data, the textures on objects, distractors and background are random images from Flickr~8k~\cite{Hodosh2013}.

\begin{figure*}
    \centering
    \begin{tabular}{ccccc}
				\raisebox{4ex}{\rotatebox[origin=c]{90}{GTA}}  &
			 \includegraphics[width=0.286\textwidth]{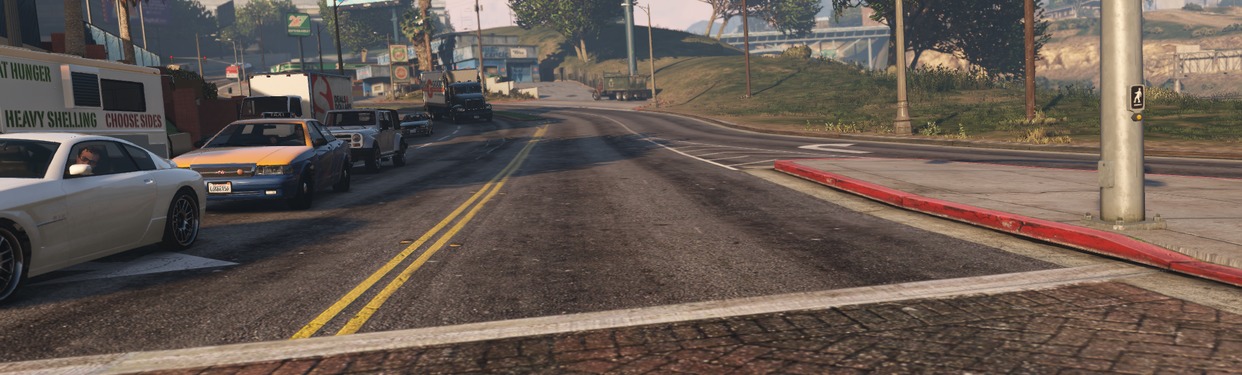} &
      \includegraphics[width=0.286\textwidth]{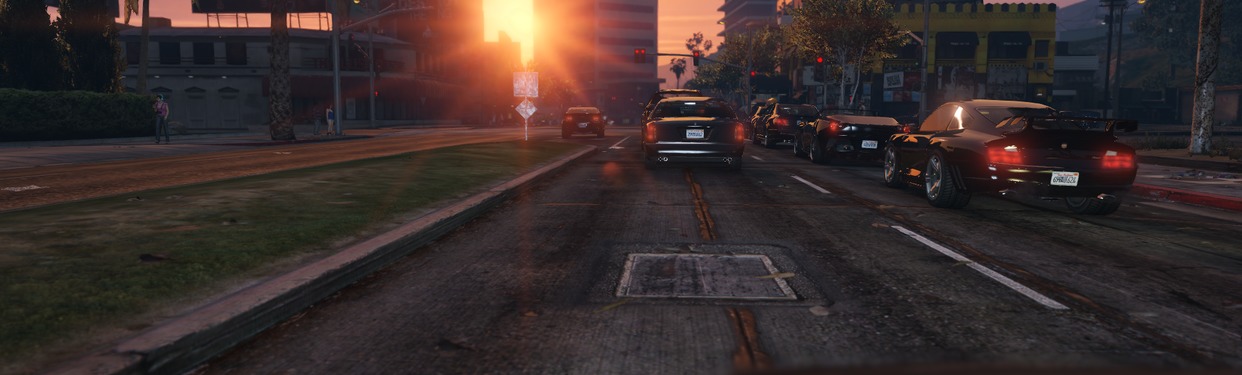} &
      \includegraphics[width=0.286\textwidth]{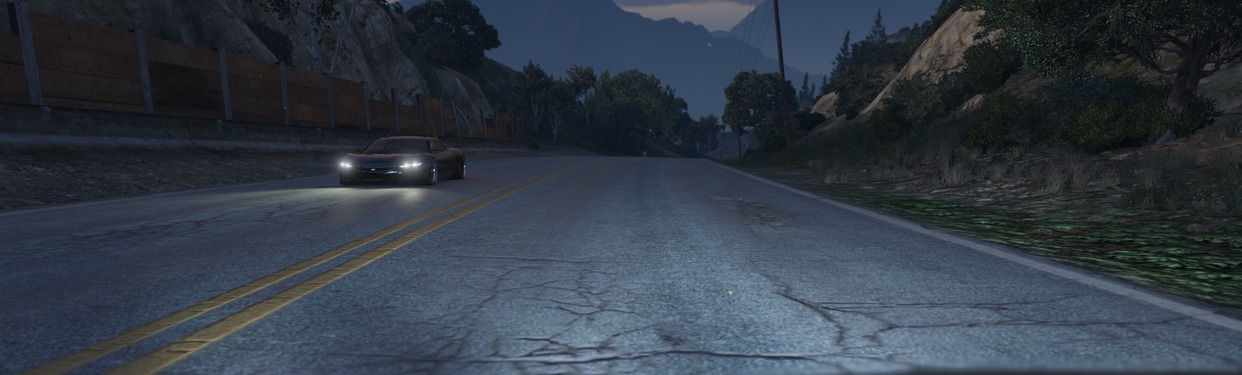} \\
			
			 \raisebox{4ex}{\rotatebox[origin=c]{90}{VKITTI}} &
      \includegraphics[width=0.286\textwidth]{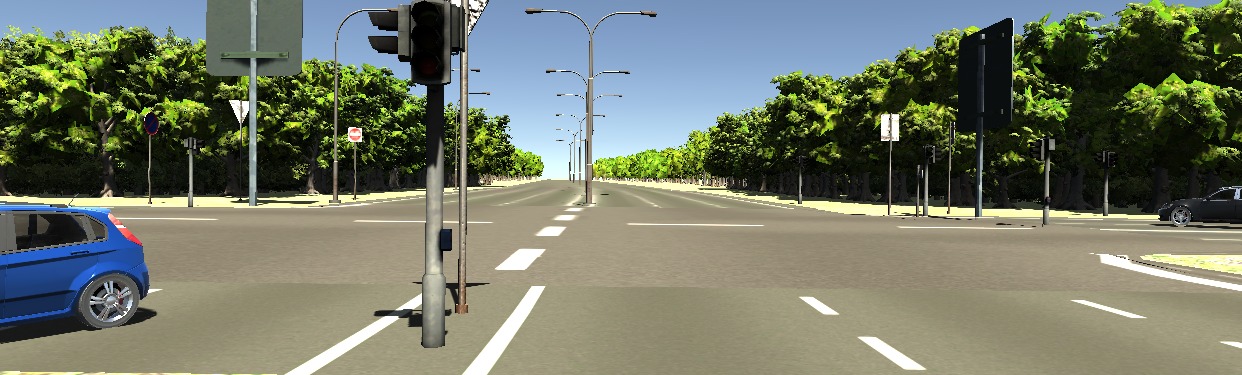} &
      \includegraphics[width=0.286\textwidth]{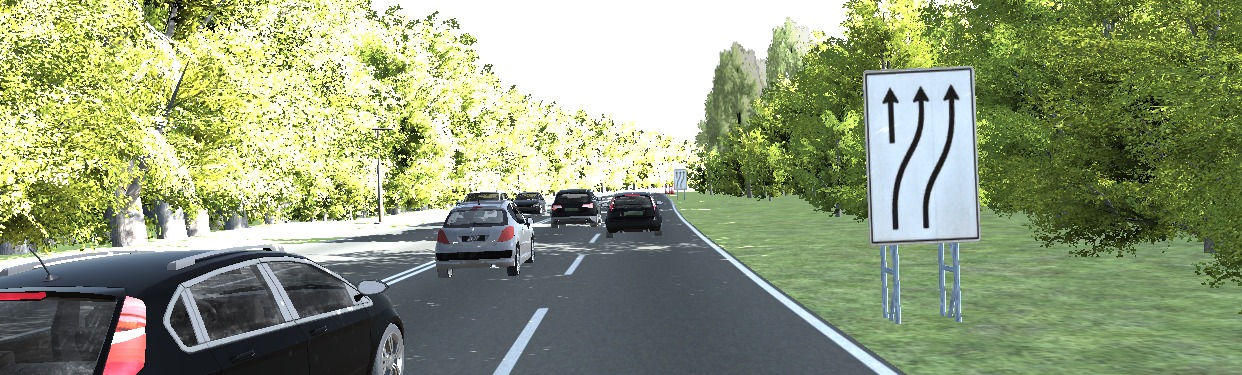} &
      \includegraphics[width=0.286\textwidth]{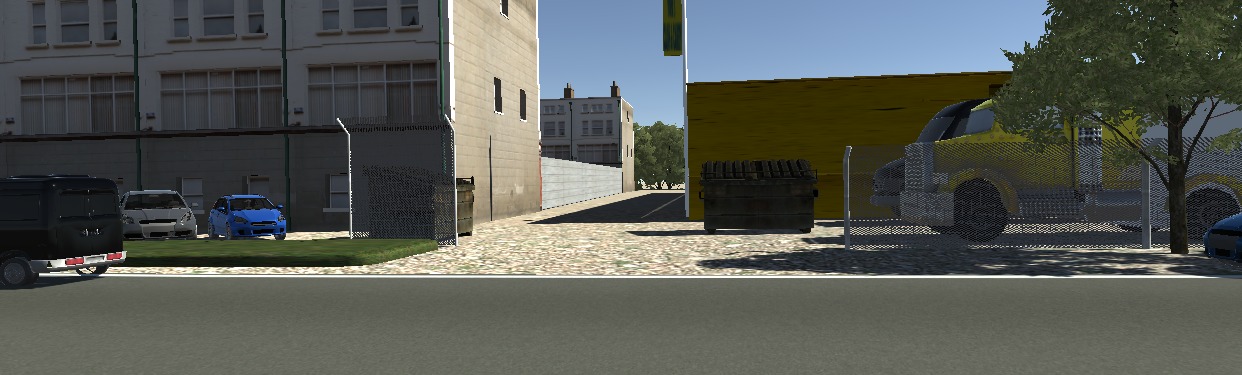} \\
			
			\raisebox{4ex}{\rotatebox[origin=c]{90}{DR}} &
			 \includegraphics[width=0.286\textwidth]{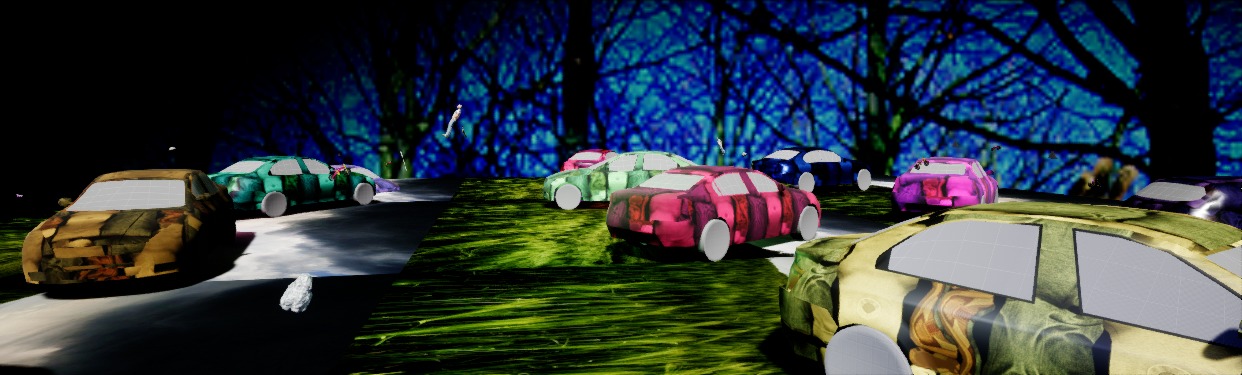} &
      \includegraphics[width=0.286\textwidth]{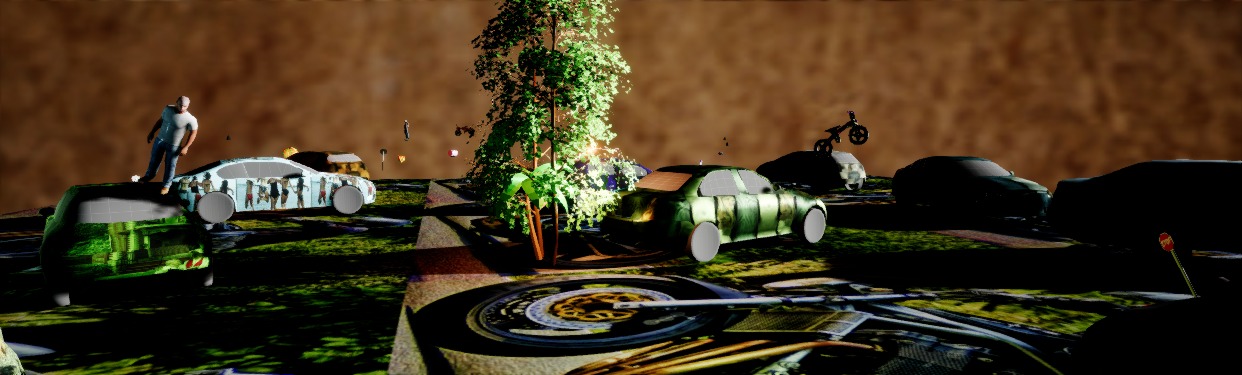} &
      \includegraphics[width=0.286\textwidth]{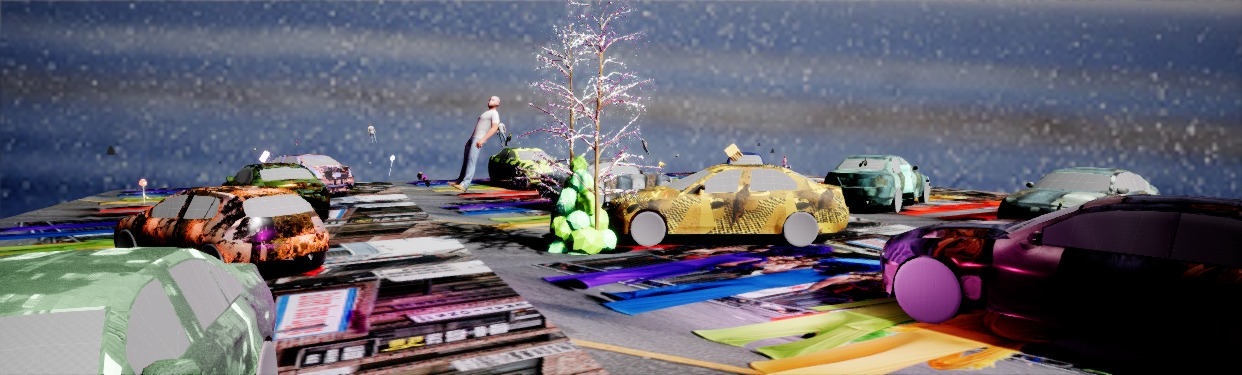} \\

			\raisebox{4ex}{\rotatebox[origin=c]{90}{SDR}} 
			\raisebox{4ex}{\rotatebox[origin=c]{90}{(ours)}} &
			\includegraphics[width=0.286\textwidth]{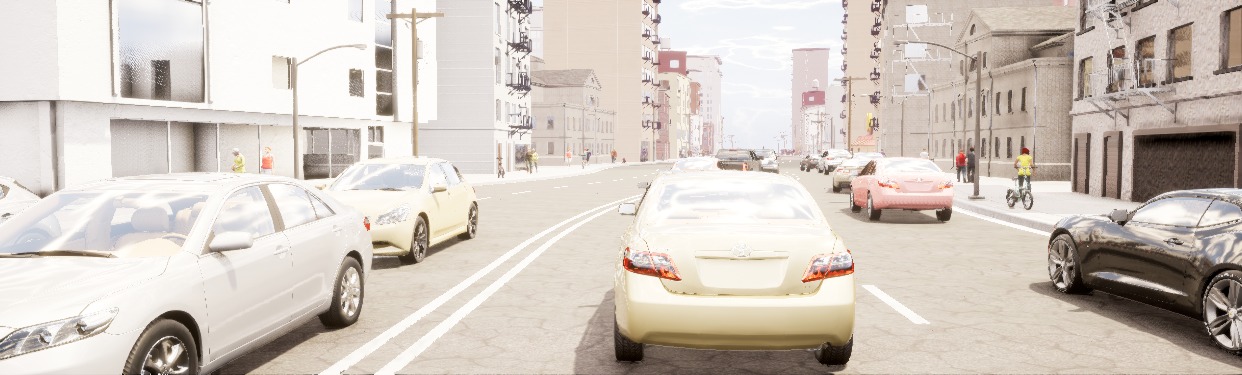} &
      \includegraphics[width=0.286\textwidth]{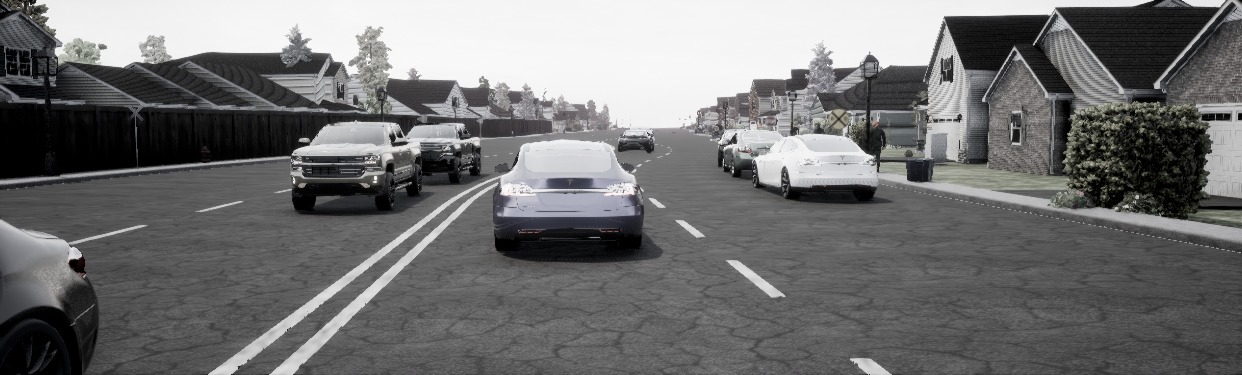} &
      \includegraphics[width=0.286\textwidth]{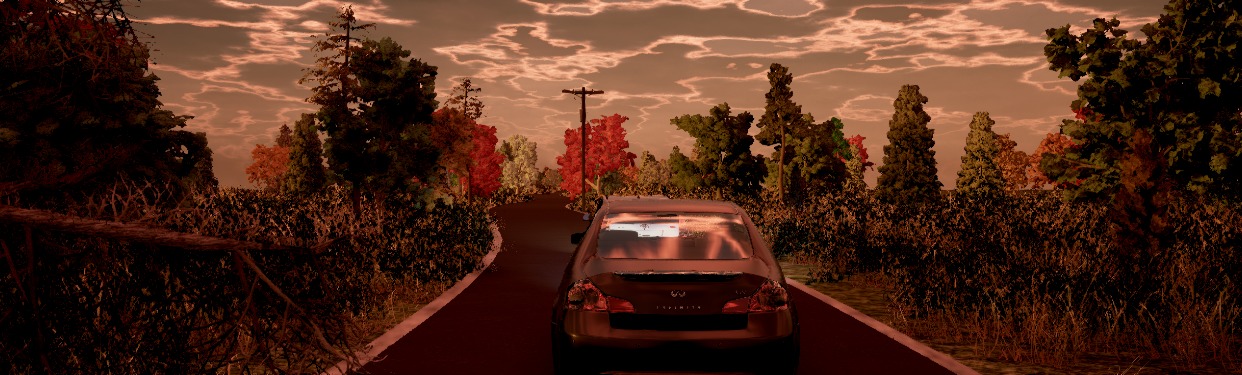} \\

    \end{tabular}
    \caption{Synthetic datasets used for training object detection models. Whereas GTA-based and Virtual KITTI both produce photo-realistic images, domain randomization (DR) intentionally avoids photorealism for variety. Structured domain randomization (SDR) strikes a balance between these two extremes, producing images that are realistic in many respects but nevertheless exhibit large variety.}
    \label{fig:synthetic-data}
\end{figure*}

Some sample images generated using this algorithm are shown in Fig.~\ref{fig:synthetic-data} along with images from other synthetic datasets used for object detection.
GTA-based data~\cite{johnson2017icra:ditmatrix,Richter_2016_ECCV,Richter_2017_ICCV} uses a large variety of assets and extensive computation time to generate realistic driving simulations, but they are not designed for object detection.
The geometry of the GTA environment is static, that is, roads, buildings, trees and foliage are always in the same positions.
SDR however can produce more variability in terms of scene geometry.
For instance it can produce countless array of road segments with varying widths and undulations, including trees, foliage and buildings that are randomly placed.
As shown in the next section, SDR outperforms the GTA-based synthetic data of Sim~200k~\cite{johnson2017icra:ditmatrix}, even though the latter uses many more assets and models, because SDR provides more variability in the geometry of the scenes as mentioned above.
VKITTI~\cite{gaidon2016CVPR} is a replica of the KITTI scenes and therefore is highly correlated with KITTI.
DR\cite{tobin17:dr,dr_tremblay_2018} generates random object placement, random object texture, random backgrounds, random distractors, and random lighting, but it lacks proper context and structure, leading to extremely non-realistic images.
In contrast, SDR uses context to place objects in realistic ways, respecting the geometry of the context boundaries, while still randomizing the position, texture, lighting, saturation, and so forth.

Sample images from various synthetic datasets for object detection are shown in Fig.~\ref{fig:synthetic-data}, along with images from DR and SDR.
GTA-based data~\cite{johnson2017icra:ditmatrix,Richter_2016_ECCV,Richter_2017_ICCV} use a large number of assets to generate realistic driving simulations.
The geometry of the GTA environment is static, so that roads, buildings, and trees are always in the same positions.
In contrast, SDR varies these parameters.
As shown in the next section, this variability is key to successful object detection.
VKITTI~\cite{gaidon2016CVPR} is a replica of the KITTI scenes and therefore is highly correlated with KITTI.
DR\cite{tobin17:dr,dr_tremblay_2018} generates random object placement, random object texture, random backgrounds, random distractors, and random lighting, but it lacks proper context and structure, leading to extremely non-realistic images.
In contrast, SDR uses context to place objects in realistic ways, respecting the geometry of the context boundaries, while still randomizing the position, texture, lighting, saturation, and so forth.

\section{Evaluation}
\label{sc:eval}

The proposed approach of SDR was evaluated for the problem of 2D bounding box detection of vehicles (cars) in the KITTI dataset~\cite{Geiger2012CVPR}.
In the following subsections we compare SDR against other approaches for generating synthetic data (\S\ref{sc:comparativestudy}), then against using real data from the same and another domain (\S\ref{sc:compare}). Afterwards we show the power of SDR as an initialization strategy (\S\ref{sc:init}), followed by an ablation study (\S\ref{sc:ablation}).

\subsection{Comparative Study}
\label{sc:comparativestudy}

We used the well-known Faster-RCNN~\cite{faster-rcnn2015NIPS} detector, which utilizes a two-stage approach.  
The first stage is a region proposal network (RPN) that generates candidate regions of interest using extracted features along with the likelihood of finding an object in each of the proposed regions.
In the second stage, features are cropped from the image using the proposed regions and fed to the remainder of the feature extractor, which predicts a probability density function over object class along with a refined class-specific bounding box for each proposal.
The architecture was trained in an end-to-end fashion using a multi-task loss.
For training, we used momentum \cite{Qian1999} with a value of 0.9, and a learning rate of 0.0003.
Resnet~V1~\cite{kaiming2016CVPR} pretrained on ImageNet~\cite{imagenet_cvpr09} was used as the feature extractor.

We trained Faster-RCNN with different synthetically-generated datasets:  Virtual KITTI (VKITTI)~\cite{gaidon2016CVPR}, Sim 200k~\cite{johnson2017icra:ditmatrix}, DR~\cite{dr_tremblay_2018}, and our SDR approach.
For DR and SDR we generated 25k images each, whereas Virtual KITTI consists of 21k images, and Sim~200k consists of 200k images.
All our experiments include standard data augmentations such as random contrast, brightness, mirror flips, and crops.

Results of detection of vehicles on the full dataset of 7500 real KITTI images are shown in Table~\ref{tbl:comparison_with_synthetic}, with the performance metric AP evaluated at 0.7 IOU overlap.
The KITTI ground truth bounding boxes are classified as Easy, Moderate, and Hard, depending upon the minimum bounding box height and maximum occlusion, with each category subsuming the previous one.  
Thus, the Hard category includes all the bounding boxes, whereas Moderate includes a subset of Hard, and Easy includes a subset of Moderate.  
From the table, it is clear that our SDR approach outperforms other synthetic datasets on all three criteria.
Although DR works well on detecting larger objects in the scene (Easy), it performs poorly on smaller objects (Moderate, Hard)
because such objects require the network to utilize context.
SDR improves upon DR by incorporating context, achieving results that are more than 2x better than DR.
Note that, although VKITTI \cite{gaidon2016CVPR} matches the distribution of the KITTI test data, it lacks the variability that SDR provides, thus causing the network to overfit to the VKITTI distribution compared with SDR.
We also trained on only the 2.2k clone videos of VKITTI, achieving noticeably worse performance due to the reduced variability.

\begin{table}
    \scriptsize
    \centering
    \begin{tabular}{p{2.0cm}>{\centering\arraybackslash}p{1.0cm}>{\centering\arraybackslash}p{1.2cm}>{\centering\arraybackslash}p{1.2cm}>{\centering\arraybackslash}p{1cm}}
        \toprule
        \scriptsize
         Dataset &  Size & \textit{Easy}  & \textit{Moderate}   &  \textit{Hard} \\
				\midrule
				 VKITTI clones~\cite{gaidon2016CVPR} & 2.2k & 49.6 & 	44.8 &	33.6 \\  %
				\midrule
				VKITTI~\cite{gaidon2016CVPR} & 21k & 70.3 & 53.6	& 39.9 \\  
        \midrule
        Sim 200k~\cite{johnson2017icra:ditmatrix} & 200k & 68.0 & 52.6 & 42.1 \\
				\midrule
        DR~\cite{dr_tremblay_2018} & 25k & 56.7 & 38.8 &	24.0 \\
				\midrule
        SDR (ours) & 25k & \textbf{77.3} & \textbf{65.6} & \textbf{52.2}\\
        \bottomrule
        \\
    \end{tabular}
    \caption{Comparison of Faster-RCNN trained on various synthetic datasets. Shown are AP@0.7 IOU for detecting vehicles on the entire real-world KITTI dataset consisting of 7500 images.}
    \label{tbl:comparison_with_synthetic}
\end{table}
\begin{figure*}
    \centering
    \begin{tabular}{cc}
      \includegraphics[width=0.48\textwidth]{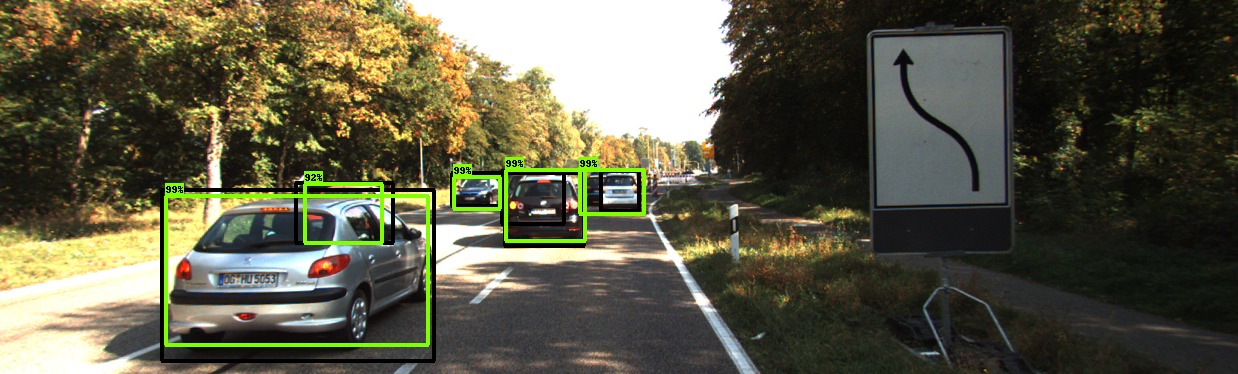} &
      \includegraphics[width=0.48\textwidth]{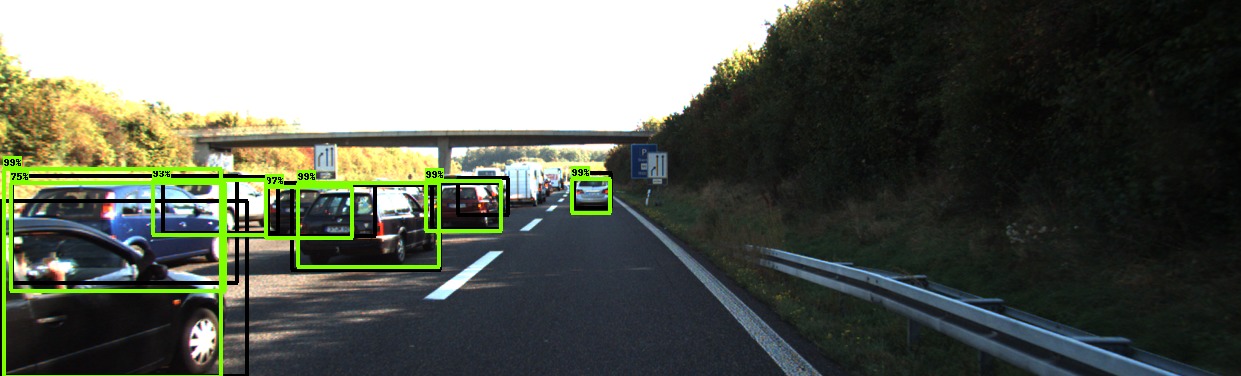} \\

      \includegraphics[width=0.48\textwidth]{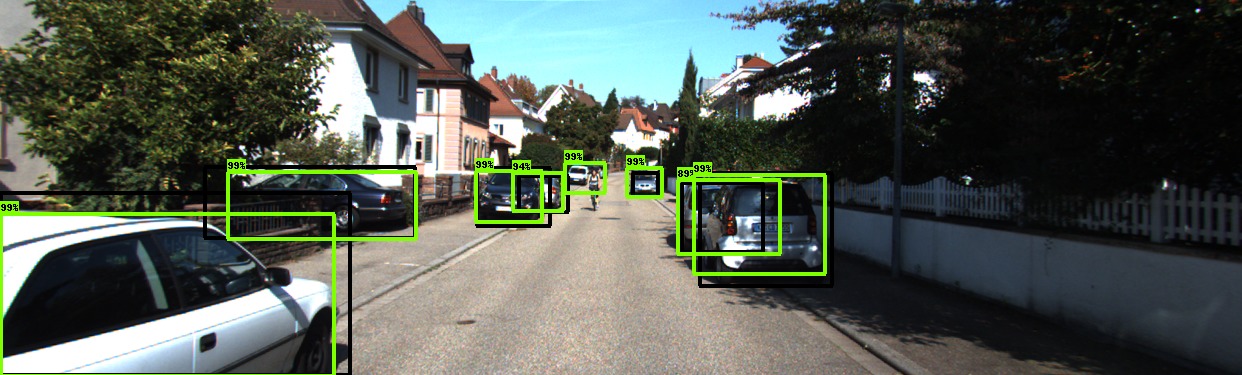} &
      \includegraphics[width=0.48\textwidth]{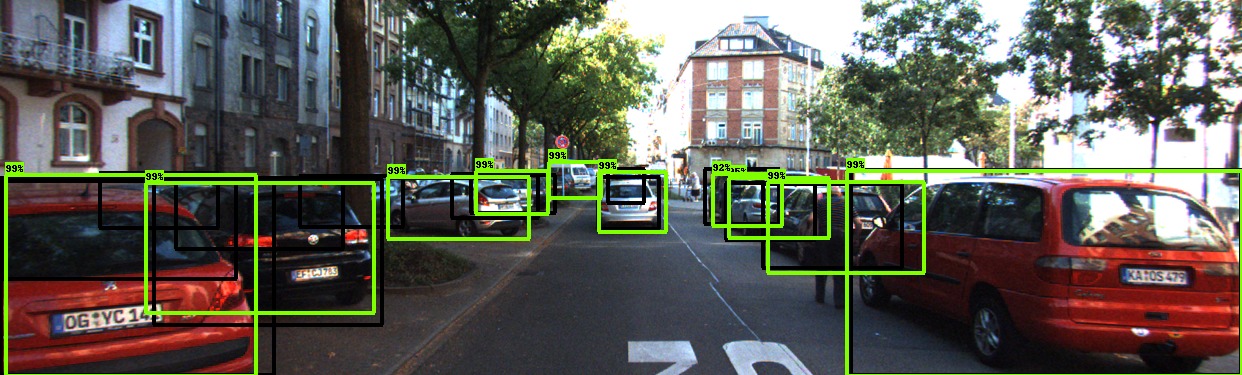} \\

    \end{tabular}
    \caption{Qualitative results on KITTI of Faster-RCNN trained only on SDR-generated synthetic data. Note the successful detection of severely occluded vehicles.  (Green boxes:  detections, black boxes:  ground truth.)}
    \label{fig:SDR_result}
\end{figure*}

Qualitative results of the detector trained on SDR are shown in Fig.~\ref{fig:SDR_result}.
Predictions are shown via green boxes, whereas ground truth is displayed via black boxes.
These results highlight the ability of an SDR-trained network to detect in complicated situations even when the network has never seen a single real KITTI image.

Table~\ref{tab:data} shows the effect of dataset size on performance. Note that performance of SDR saturates quickly around 10k images and that with just 1000 images we already achieve 43.7 AP. In contrast, the performance of DR does not saturate until approximately 50k images.

\begin{table}
    \centering
	\begin{tabular}{ccccccc}
			\toprule
		   Dataset size & 1k & 2.2k & 10k & 25k &50k & 100k \\
			\midrule
			 DR & 20.6 & 22.1 & 23.2 & 24.0 & 25.8 & 25.6 \\
			\midrule
       SDR & \textbf{43.7} & \textbf{46.0} & \textbf{51.9} & \textbf{52.5} & \textbf{51.1} & \textbf{51.6} \\
			\bottomrule
			\\
\end{tabular}
\caption{Effect of dataset size on AP at 0.7 IOU for detection of vehicles on KITTI Hard by DR and SDR when evaluated on a subset of real KITTI images. Performance saturates around 10k for SDR and 50k for DR.}
\label{tab:data}
\end{table}

\subsection{Domain Gap}
\label{sc:compare}

To compare training using synthetic versus real data, we evaluate using Faster-RCNN~\cite{faster-rcnn2015NIPS} on a subset of 1500 randomly selected real KITTI images, allowing us to use the remaining 6k real KITTI images for training.  
The results are shown in Table~\ref{tbl:comparison_with_real}.
Since the distributions of KITTI training and test images match each other, it is difficult for a network trained only on synthetic data to compete.
Nevertheless, these results highlight that there is not only a reality gap between synthetic and real data, but there are also significant domain gaps between various real-world datasets.
These gaps are evident by the relatively poor performance of networks trained on real data from a different domain (BDD100K) but tested on KITTI.
Significantly, SDR outperforms this real dataset.

\begin{table}
    \scriptsize
    \centering
    \begin{tabular}{p{1.2cm}>{\centering\arraybackslash}p{0.8cm}>{\centering\arraybackslash}p{0.8cm}>{\centering\arraybackslash}p{1.1cm}>{\centering\arraybackslash}p{1.1cm}>{\centering\arraybackslash}p{1cm}}
        \toprule
        \scriptsize
         Dataset &  Type & Size & \textit{Easy}  & \textit{Moderate}   &  \textit{Hard} \\
        \midrule
        DR~\cite{dr_tremblay_2018} & synth & 25k & 56.8 &	38.0	& 23.9 \\
				\midrule
        SDR (ours) & synth & 25k & 69.6 & 65.8 & 52.5\\
				\midrule
				\midrule
				BDD100K~\cite{bdd100k} & real & 70k & 59.7 & 54.3 & 45.6 \\
				\midrule
				KITTI  & real & 6k & \textbf{85.1} &	\textbf{88.3} & \textbf{88.8} \\
				\bottomrule
        \\
    \end{tabular}
    \caption{Comparison of Faster-RCNN trained on synthetic data (DR, SDR) or real data (BDD100K, KITTI). Shown are AP@0.7 IOU for vehicle detection from a subset of 1500 images from the real-world KITTI dataset.  Although it is difficult for synthetic data to outperform real data from the same distribution as the test set (KITTI), our SDR approach nevertheless outperforms real data from other distribution (BDD100K).}
    \label{tbl:comparison_with_real}
\end{table}

\subsection{SDR as an Initialization Strategy}
\label{sc:init}

\begin{figure}
	\begin{center}
        \includegraphics[width=1\linewidth]{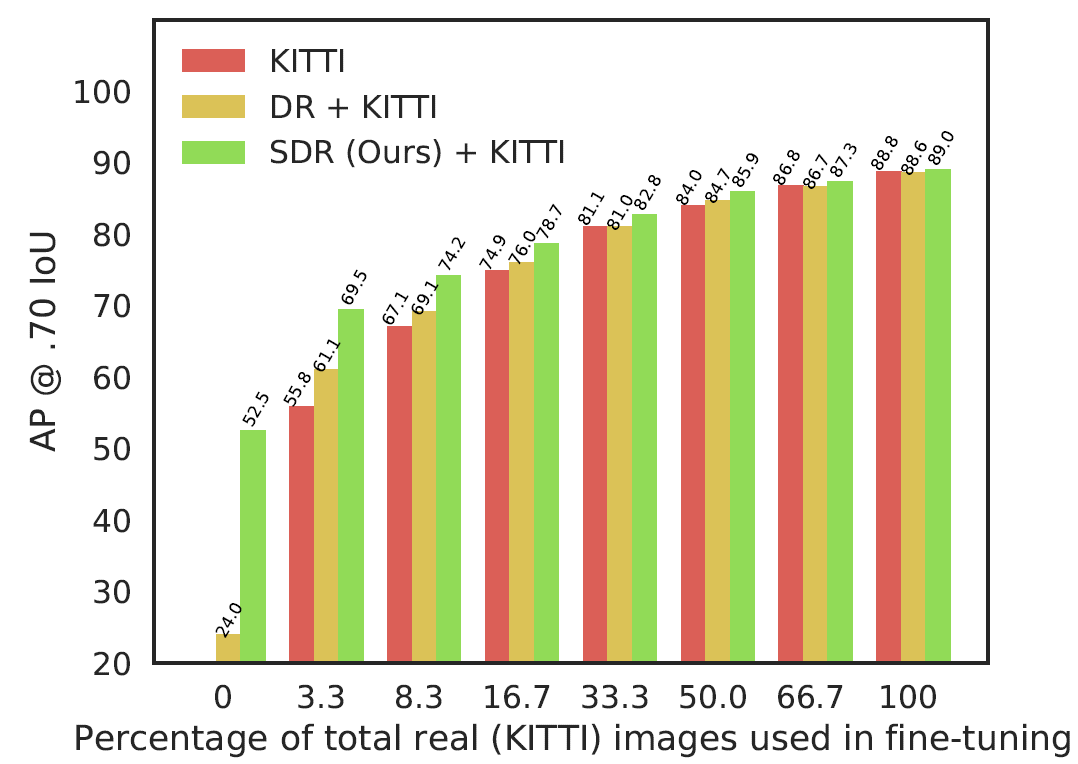}
	\end{center}
   \caption{The performance of real data is boosted when pretrained with SDR-generated synthetic data.  Improvement is especially pronounced when only a few labeled real images are available.}
	\label{fig:init}
\end{figure}

SDR is also a good way to initialize a network when an insufficient amount of labeled real data is available.
The results of the initialization/fine-tuning experiment are shown in Fig.~\ref{fig:init}.
Training solely using SDR yields an AP of 52.5 on the subset of 1500 real KITTI images, as mentioned above.
We then fine-tune using some number of real KITTI images (that is, some percentage of the remaining 6000 images), using momentum \cite{Qian1999} with a value of 0.9, and a learning rate of 0.0003.
For comparison, we also train on the same number of real KITTI images using the same learning rate.
As seen in the figure, the performance of SDR+real KITTI is always higher than KITTI alone, showing the importance of using SDR for initializing a network even when real labeled data is available for training.
The performance of SDR+real is also better than DR+real, especially for smaller labeled datasets.

\subsection{Ablation Study}
\label{sc:ablation}

\begin{figure}
\begin{center}
\includegraphics[width=1\linewidth]{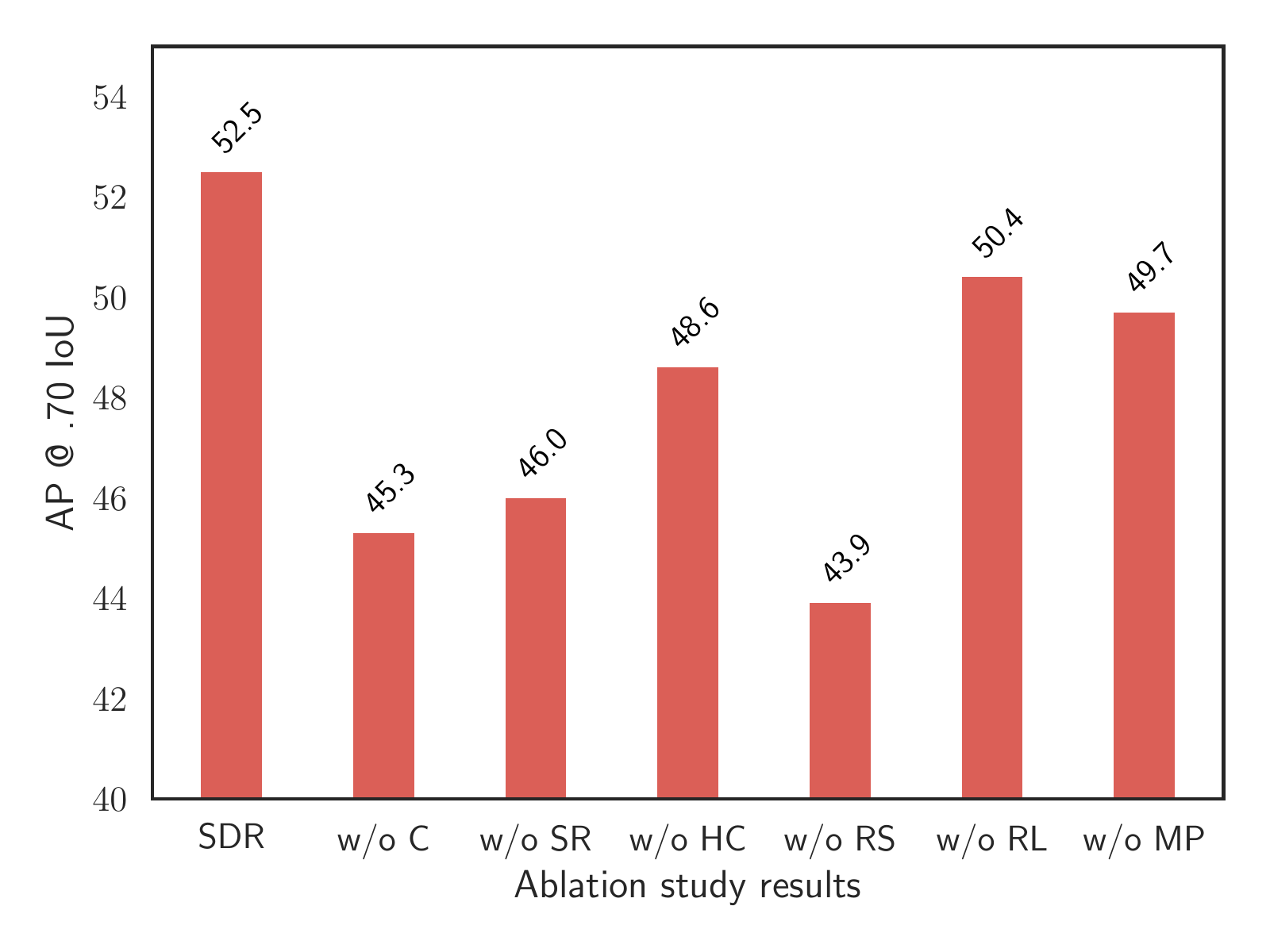}
\end{center}
\caption{Ablation study for full SDR, without context (C), without scene randomization (SR), without high contrast (HC), without random saturation (RS), without random light (RL) and without multiple pose (MP).}
\label{fig:ablation}
\end{figure}

In this section we present the effects of individual SDR parameters.
For this experiment we used the same Faster-RCNN~\cite{faster-rcnn2015NIPS} network and Resnet V1~\cite{kaiming2016CVPR} pretrained on ImageNet \cite{imagenet_cvpr09} as feature extractor and the same validation set as Sec.~\ref{sc:compare}.
Our previous DR parameter study~\cite{dr_tremblay_2018} showed lighting to be most important parameter.
For SDR we find other parameters (e.g., context, saturation and contrast) to be more important, as shown in Fig.~\ref{fig:ablation}.

The details of the ablation study are as follows.  \textbf{C:} Instead of roads, sidewalks, trees, and other 3D objects placed in the scene, random 2D background images were used.  This result shows the importance of context. \textbf{SR:} Instead of a variety of scenarios, images were generated from only rural (46.0 AP), suburban (47.7 AP), or urban (51.9 AP) scenes.  This result reveals the importance of variety in the scenes.  \textbf{HC:}  For SDR, contrast is fixed at 150\% of normal; for this experiment, contrast was set to 100\%.  \textbf{RS:} Random saturation was removed.  This change has the largest effect, suggesting the importance of the texture gap between real and synthetic data.  \textbf{RL:} Lighting was fixed to a single time of day (broad daylight).  \textbf{MP:} Vehicle poses were always fixed to be within a lane, thus degrading performance when detecting parked vehicles or vehicles on side streets.

\section{Conclusion}
We have introduced structured domain randomization (SDR), which imposes structure onto domain randomization (DR) in order to provide context. For detecting vehicles, for example, SDR places the vehicles on roads, thus enabling the neural network during training to learn the relationship between them. Through experiments we show that this improves performance significantly over DR. In this paper we have shown that SDR achieves state-of-the-art results for vehicle detection on the KITTI dataset compared with both other synthetically-generated data as well as real-world data from a different domain.  We have also shown that pretraining on SDR improves results from real data.  In future research, we intend to study SDR for detecting multiple object classes, semantic segmentation, instance segmentation, and other computer vision problems.

\section*{Acknowledgments}

The authors thank Sean Taylor, Felipe Alves, and Liila Torabi for their help with the project.

{\small
\bibliographystyle{IEEEtran}
\bibliography{main}
}

\end{document}